\documentclass[10pt,twocolumn,letterpaper]{article}

\usepackage[pagenumbers]{cvpr} % To force page numbers, e.g. for an arXiv version

\definecolor{MyPurple}{HTML}{AF58BA}

\usepackage[T1]{fontenc} 
\newcommand{\modelname}{{\scshape OpenHelix}\xspace}

\definecolor{cvprblue}{rgb}{0.21,0.49,0.74}
\usepackage[pagebackref,breaklinks,colorlinks,allcolors=cvprblue]{hyperref}

\def\confName{CVPR}
\def\confYear{2025}

\usepackage{algorithm}
\usepackage{algpseudocode}
\usepackage{booktabs}       
\usepackage{amsfonts}       
\usepackage{nicefrac}       
\usepackage{microtype}      
\usepackage{xcolor}         
\usepackage{wrapfig}
\usepackage{enumitem}
\usepackage{graphicx}
\usepackage{float}
\usepackage{amsmath}
\usepackage{amssymb}
\usepackage{array}
\usepackage{multicol,multirow}
\usepackage{setspace}
\usepackage{makecell}
\usepackage{tasks}
\usepackage{pifont}
\usepackage{colortbl}
\usepackage{titletoc}
\usepackage{color}
\usepackage{caption, subcaption, overpic, textpos}
\usepackage{natbib}
\usepackage{url}
\newcommand{\act}{{\texttt{<ACT>}}}
\usepackage{multirow}
\usepackage{graphicx}
\usepackage{booktabs}
\usepackage{adjustbox} % 用于调整表格大小

\definecolor{baselinecolor}{gray}{.9}

\title{\modelname: A Short Survey, Empirical Analysis, and Open-Source Dual-System VLA Model for Robotic Manipulation}

\author{Can Cui$^{1}$,
Pengxiang Ding$^{12}$\thanks{\ Project Lead. {Email: dingpx2015@gmail.com}}\hspace{0.35em}, 
Wenxuan Song$^{4}$,
Shuanghao Bai$^{3}$,\\
Xinyang Tong$^{1}$,
Zirui Ge$^{2}$, 
Runze Suo$^{1}$,
Wanqi Zhou$^{3}$,
Yang Liu$^{1}$,
Bofang Jia$^{1}$,\\
Hangyu Liu$^{12}$,
Mingyang Sun$^{12}$,
Han Zhao$^{12}$,
Siteng Huang$^{1}$,
Donglin Wang$^{1}$\\
$^{1}$Westlake University \quad 
$^{2}$Zhejiang University \quad 
$^{3}$Xi'an Jiaotong University \quad
$^{4}$HKUST(GZ) \quad \\
}

\begin{document}
\maketitle
\begin{abstract}

Dual-system VLA (Vision-Language-Action) architectures have become a hot topic in embodied intelligence research, but there is a lack of sufficient open-source work for further performance analysis and optimization. To address this problem, this paper will summarize and compare the structural designs of existing dual-system architectures, and conduct systematic empirical evaluations on the core design elements of existing dual-system architectures. Ultimately, it will provide a low-cost open-source model for further exploration. Of course, this project will continue to update with more experimental conclusions and open-source models with improved performance for everyone to choose from. Project page: https://openhelix-robot.github.io/.

\end{abstract}    

\section{Short Survey}
\label{sec:intro}

\subsection{Definition of VLAs}
Traditional policy learning has primarily focused on training novel behaviors from scratch using lightweight models. These models demonstrate high sensitivity to environmental perturbations (both visual and textual) and exhibit limited generalization capabilities. However, with the emergence of large language models (LLMs) and vision-language models (MLLMs), this landscape is undergoing significant transformation. These models, trained on Internet-scale data with vast parameter spaces, have demonstrated exceptional proficiency in text generation and visual comprehension, consequently generating substantial interest in their application to robotics policy training.
In this context, RT-2 introduced the pioneering concept of vision-language-action models (VLAs), which co-fine-tune state-of-the-art vision-language models on both robotic trajectory data and Internet-scale vision-language tasks. VLAs have demonstrated remarkable improvements in generalizing to novel objects and semantically diverse instructions while exhibiting a range of emergent capabilities. Furthermore, VLAs possess the potential to revolutionize robotic skill acquisition methodologies, as they serve as powerful foundation models that can be directly fine-tuned to adapt to domain-specific robotic applications. Subsequently, research on VLAs has proliferated rapidly, offering a highly promising approach to addressing the challenges associated with robotic deployment.

\subsection{Limitation of VLAs}
Directly applying VLAs in domain-specific or real-world scenarios remains challenging due to the following limitations:
(1) The large and cumbersome model size of VLAs makes achieving efficient real-time performance difficult. For instance, RT-2 demonstrated that their 55B model operates at 1 to 3 Hz, while the 5B model runs at around 5 Hz under experimental conditions. In contrast, traditional lightweight models, such as BC-Transformer, operate at much higher speeds (around 50 Hz).
(2) Pre-training is resource-intensive, and end-to-end fine-tuning of pre-trained VLAs on embodied data is also challenging due to domain shift and catastrophic forgetting.
Leveraging existing MLLMs and VLAs for practical applications, while retaining their remarkable capabilities in multimodal understanding, reasoning, and generation, and ensuring fast inference for coherent actions, remains a challenge that needs to be addressed.

\begin{table*}[ht]
\centering
\caption{
\textbf{Methods Comparisons of Dual-System VLAs}. 
Here, L, R, P, D, T, and PC represent different modalities: Language, RGB, Proprioception, Depth, Tactile, and Point Cloud, respectively. 
FT denotes fine-tuning.
Pretrain and Scratch denote fine-tuning a pre-trained policy head and training a policy head from scratch, respectively.
}
\label{tab:survey}
\resizebox{\textwidth}{!}{
\begin{tabular}{ccccccccc}
\toprule
\multirow{2}{*}{Method} & \multicolumn{3}{c}{System 2}  & \multirow{2}{*}{Latent Rep.} & \multicolumn{3}{c}{System1}  \\
\cmidrule(r){2-4} \cmidrule(r){6-8}
& Model & Input & Training &  & Policy Head   & Sensory & Training \\
\midrule
LCB~\cite{shentu2024llms}    & LLaVA-7B  & L+R & Lora FT & Lang(<ACT>)  & 3D Diffusion Actor~\cite{ke20243d}  & {R+P+PC}    & Pretrain \\
DP-VLA~\cite{han2024dual} & OpenVLA-7B  & L+R & Frozen  & Vis+Lang    & Transformer & R+P     & Scratch  \\
HiRT~\cite{zhang2024hirt}   & InstructBLIP-7B & L+R & Lora FT & MaxPooling(Vis+Lang) & RT-1~\cite{brohan2023rt1roboticstransformerrealworld} & R & Scratch  \\   
Robodual~\cite{bu2024towards} & OpenVLA-7B & L+R & Lora FT & Action+Lang & DiT & R+D+T+P & Scratch  \\
DexVLA~\cite{wen2025dexvla} & Qwen2-VL-2B & L+R & Lora FT & Lang & ScaledDP~\cite{zhu2024scaling} & {R+P} & Scratch  \\
Helix & N/A   & L+ R + P & N/A & N/A  & Transformer   & R+P  & N/A \\
\bottomrule
\end{tabular}
}
\end{table*}

\subsection{Definition of Dual-System VLAs}
As a result, the Dual-System VLAs were introduced. LCB~\cite{shentu2024llms} pioneered the adoption of the Dual-System VLA structure, while DP-VLA is the first to incorporate \textbf{dual-process theory} to provide a personified explanation for the rationale underlying this architecture.
Dual-process theory~\cite{tversky1974judgment, kahneman2011thinking, evans2008dual, neys2006dual} conceptualizes human cognition as operating through two distinct systems:

\noindent \textbf{1. System 1 is fast, automatic, intuitive, and unconscious.} It operates effortlessly and relies on heuristics to make judgments and decisions. System 1 is responsible for our immediate reactions, such as making simple or routine choices. System 1 often leads to biases and errors because it relies on mental shortcuts, such as heuristics, which can be effective in some situations but also result in systematic mistakes. 
    In the robotics domain, this system closely resembles traditional lightweight policy networks, which are efficient but often task-specific.
    
\noindent \textbf{2. System 2 is slow, deliberate, effortful, and conscious.} It involves reasoning, logic, and careful evaluation of evidence. System 2 is engaged when performing cognitively demanding tasks, such as solving complex problems or making thoughtful decisions. 
    System 2, while generally more accurate, requires greater cognitive resources and is also prone to errors when cognitive load is high or attention is limited.
    In robotics, this system is analogous to large-scale models like MLLMs and VLAs, which are computationally heavy but offer superior generalization capabilities.

\noindent \textbf{3. Although the two systems operate in parallel, they update information at different frequencies. }The slower System 2-like component updates less frequently and is responsible for making more deliberate decisions based on high-level representations. In contrast, the faster System 1-like component updates at a higher frequency to rapidly generate the low-level actions required for real-time robotic control. Notably, the information from the slow system is subject to temporal delay.
This architecture addresses the aforementioned challenge by simultaneously enabling efficient real-time inference while preserving the multimodal reasoning capabilities of large models.

\subsection{Current Dual-System VLAs}
We introduce recent Dual-System VLA approaches below, with a comparative analysis of their distinctive features summarized in Table~\ref{tab:survey}. \textbf{It is important to note that for synchronous inference to occur, System1 must incorporate real-time perception inputs (such as RGB images).} According to this criterion, approaches like $\pi_0$~\cite{black2024pi_0}, GR00TN1~\cite{bjorck2025gr00t}, and similar methodologies cannot be properly classified within the dual-system framework as they lack this essential characteristic.

\noindent \textbf{LCB} adopts LLaVA as its System 2. Given a high-level task description and an RGB observation, LLaVA generates a textual action description along with an <ACT> token. The <ACT> token, derived from the final layer, serves as a high-level latent goal. System 1 is a pre-trained 3D Diffusion Actor that takes the RGB image, point cloud, and <ACT> token as input to generate actions. System 2 is fine-tuned using LoRA, while System 1 is fine-tuned in a standard manner.

\noindent \textbf{DP-VLA} introduces dual-process theory to justify the rationale behind the dual-system architecture. It presents a more generalizable design choice, where System 2 is not limited to MLLMs, but can also be VLAs that are pre-trained on robot data. In experiments, DP-VLA adopts OpenVLA as System 2 and uses its encoder to extract latent representations from language instructions and RGB observations to guide System 1. System 1 is implemented using a Transformer architecture, which encodes RGB images and proprioceptive inputs into actions. System 2 is kept frozen, while System 1 is trained from scratch.

\noindent \textbf{HiRT} adopts InstructBLIP as System 2 and utilizes the final-layer representations obtained from encoding both language instructions and RGB observations. These representations are processed with MAP pooling to produce MLLM latent features that guide System 1. System 1 uses an EfficientNet-B3 backbone combined with a MAP block to encode RGB inputs into actions. System 2 is fine-tuned using LoRA, while System 1 is trained from scratch.

\noindent \textbf{Robodual} adopts OpenVLA as System 2 and extracts latent representations from language instructions and RGB observations. It uses both the task latent derived from the instruction and the final action latent as guidance signals. System 1 encodes RGB, depth, tactile, and proprioceptive inputs using a ViT-based encoder, and employs a Perceiver Resampler to distill key features. A DiT model then generates actions by conditioning on the distilled features, the task latent, and a noisy action input. System 2 is fine-tuned using LoRA, while System 1 is trained from scratch.

\begin{figure*}[t]
\begin{center}
\vspace{-1.2em}
  \includegraphics[width=1.0\textwidth]{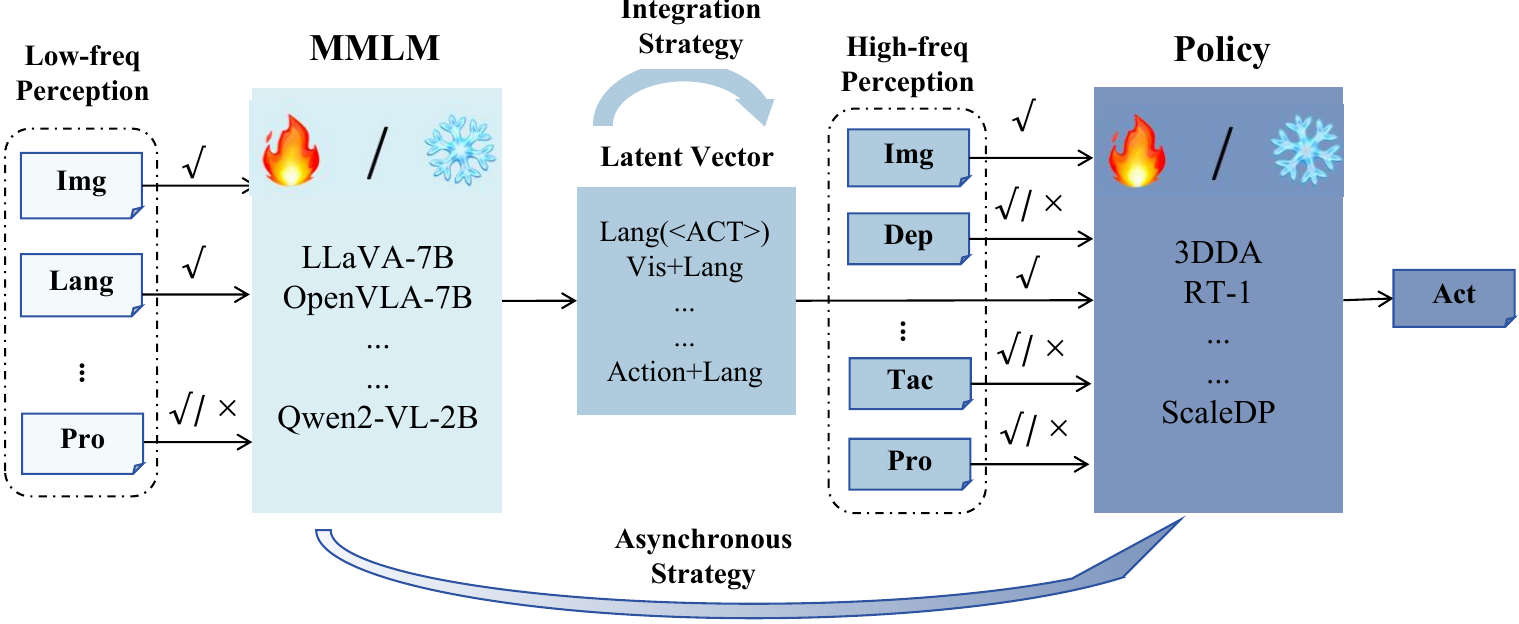} 
\end{center}
\caption{\textbf{Key Design of Dual-System VLAs.}
It mainly includes: MMLM Selection, Policy Selection, Latent Feature Representation Selection, MLLM Training Strategy, Policy Training Strategy, Dual-System Integration Strategy, and Dual-System Asynchronous Strategy.}
\label{newfig: overview}
\end{figure*}

\subsection{Key Design of Dual-System VLAs}
The key question lies in how to design the architecture of these two systems and structure the information flow from the slower system to the faster one in a way that preserves the strengths of the System 2-like component while effectively guiding the System 1-like component to execute robotic actions. Achieving this delicate balance is essential for building robotic systems that are both highly performant and generalizable. As shown in Figure~\ref{newfig: overview}, to achieve this objective, several core issues need to be addressed:

\noindent \textbf{1. MLLM Selection.}
For different VLA scenarios, requirements of MLLMs vary. For building a model suitable for robotic scenarios, the MLLM model should be selected appropriately. For example, Flower~\cite{reussflower}'s foundation model has strong capabilities in spatial awareness/low-level vision, therefore achieving current SOTA across various tasks; MiniVLA~\cite{belkhale2024minivla} chose Qwen-VL 0.25B as its foundation model to reduce model inference costs and burden. Therefore, in an era of rapidly evolving MLLMs, we should clarify what kind of MLLM model is lightweight enough yet sufficient to complete robotic tasks, which is a problem that needs to be addressed.

Furthermore, whether a MLLM pre-trained on robotic data is necessary remains an unresolved question. Training on extensive robotic datasets not only reduces the domain gap, but also, by exposing the model to more language instructions, makes its performance exceptionally robust on language instruction following tasks, as demonstrated in experiments with Robodual~\cite{bu2024towards}.

\noindent \textbf{2. Policy Selection.}
The choice of small models is relatively less controversial, with the current general consensus being that models based on DiT structure and Flow Matching structure can both meet current needs. However, with the introduction of new policy models such as CARP~\cite{gong2024carp}, Dense Policy~\cite{su2025dense}, and other new architectures, downstream small models may also see new designs.
Additionally, like Robodual~\cite{bu2024towards}, whether downstream small models need more modal information, and which modal information is essential for system1, is also a potential question.

\noindent \textbf{3. Latent Feature Representation Selection.}
For the selection of latent tokens, this is the most complex aspect of dual-system tasks that urgently needs research. Previous methods have shown significant differences in their approaches. We need to consider not only dual-system work but also single-system work such as  ~\cite{black2024pi_0,bjorck2025gr00t,li2023vision}
For DP-VLA~\cite{han2024dual}, they directly chose the last layer hidden embedding of the MLLM large model. Meanwhile, GR00T-N1~\cite{bjorck2025gr00t} selected hidden embeddings from middle layers, considering that middle-layer features might contain more visual information and could reduce inference time. Taking this further, Roboflamnigo~\cite{li2023vision} and HiRT~\cite{zhang2024hirt} used maxpooling of the last layer language features and visual features as downstream conditions.
Beyond directly utilizing MLLM hidden embeddings, some models (e.g., LCB~\cite{shentu2024llms}) additionally introduced the concept of <ACT> tokens, hoping to bridge upstream and downstream through fine-tuning a special token, which showed promising results.
The above two approaches were further developed in Robodual~\cite{bu2024towards}, which adopted multiple <ACT> tokens while also incorporating last-layer language features as latent feature representations.
Of course, beyond the robotics domain, there are more ingenious works utilizing hidden states, such as Metaquery~\cite{pan2025transfer} and LEGO~\cite{lai2024lego}, which employed more sophisticated methods for latent feature selection.
In summary, the selection of latent feature representations will be an important research focus for dual-system models, exploring more suitable latent features for downstream action generation models.

\noindent \textbf{4. MLLM Training Strategy.}
Regarding how to train MLLMs, the main consideration is examining whether we can maintain the model's generalization capabilities without loss while also ensuring good integration with downstream tasks. Currently, the main approaches include frozen and fine-tuning methods, but exploring whether there are better fine-tuning techniques remains a valuable direction for research.

\noindent \textbf{5. Policy Training Strategy.}
Regarding how to train the Policy, the main consideration is whether to reduce the model's training cost. If we can take a pre-trained policy and fine-tune it, this could greatly reduce the overall training time. Of course, if we train from scratch, whether the different optimization objectives would make model convergence difficult is also an unknown factor that needs to be explored.

\noindent \textbf{6. Dual-System Integration Strategy.}
Regarding Integration strategies, the main focus is how to embed latent information as a condition into downstream models.
In LCB~\cite{shentu2024llms}, the authors demonstrated using CLIP loss to constrain upstream latent features to be similar to the original text CLIP embedding to connect upstream and downstream components. However, this approach clearly limits the model to only handle cases it was trained on downstream, negating the purpose of introducing the generalization capabilities of MLLM models.
Additionally, when introducing a new embedding, differences in dimensions between upstream and downstream models are inevitable, making it common to add a projector between them. However, how to train this projector requires careful consideration. In subsequent experiments, when the downstream policy is a pre-trained one, it becomes critically important to pre-align the projector without training the MLLM. If both are unfrozen and trained simultaneously, the model training will collapse.
Therefore, the Dual-System Integration Strategy is a crucial aspect.

\noindent \textbf{7. Dual-System Asynchronous Strategy.}
Lastly, there are asynchronous strategies for dual-system models. LCB~\cite{shentu2024llms}, HiRT~\cite{zhang2024hirt}, and Robodual~\cite{bu2024towards} employ different asynchronous approaches, with LCB~\cite{shentu2024llms} being the most naive, using synchronous training but asynchronous testing.
Theoretically, differences in inference frequency between upstream and downstream components could affect final performance. However, this is not entirely accurate - if the upstream features being provided aren't effective to begin with, perhaps asynchronous inference between upper and lower layers is merely a pseudo-requirement.
Therefore, more experiments are needed to verify this.

\section{Empirical Evaluations}
According to the above survey, it is evident that current dual-system models exhibit substantial variation across multiple dimensions, including the choice of base vision-language model (MLLM), downstream policy architecture, and latent selection mechanisms~\cite{shentu2024llms, han2024dual, zhang2024hirt, bu2024towards, wen2025dexvla}. These discrepancies highlight the urgent need for a systematic and fair comparison, in order to assess the rationale behind different design choices and to establish a reference framework for future model development.

In this work, we standardize experimental conditions 1, 2, 3, and 7 to ensure consistency, and focus our evaluation primarily on conditions 4, 5, and 6. These conditions involve widely applicable techniques that are largely independent of the specific choices made in conditions 1, 2, 3, and 7. Through this controlled comparison, we aim to offer insights that may inspire and guide future research in this domain.
We plan to extend our evaluation to cover additional conditions in future work. All updates and ongoing developments will be made available in our official GitHub repository: https://github.com/OpenHelix-robot/OpenHelix/.

\subsection{Experiment setup}

\noindent\textbf{Model Selection.}
To ensure consistency with LCB~\cite{shentu2024llms}, we adopt LLaVA1.0~\cite{liu2023visual} as the visual language model (MLLM) throughout this paper.
To eliminate discrepancies caused by differing policy architectures, all subsequent experiments utilize 3DDA~\cite{ke20243d} as the unified downstream policy.
The integration of latent representations is implemented in accordance with the methodology introduced in LCB~\cite{shentu2024llms}.
In line with LCB~\cite{shentu2024llms}, we employ synchronous training and asynchronous testing for experiments involving asynchronous settings.

\begin{figure}[!t]
\begin{center}
  \includegraphics[width=0.48\textwidth]{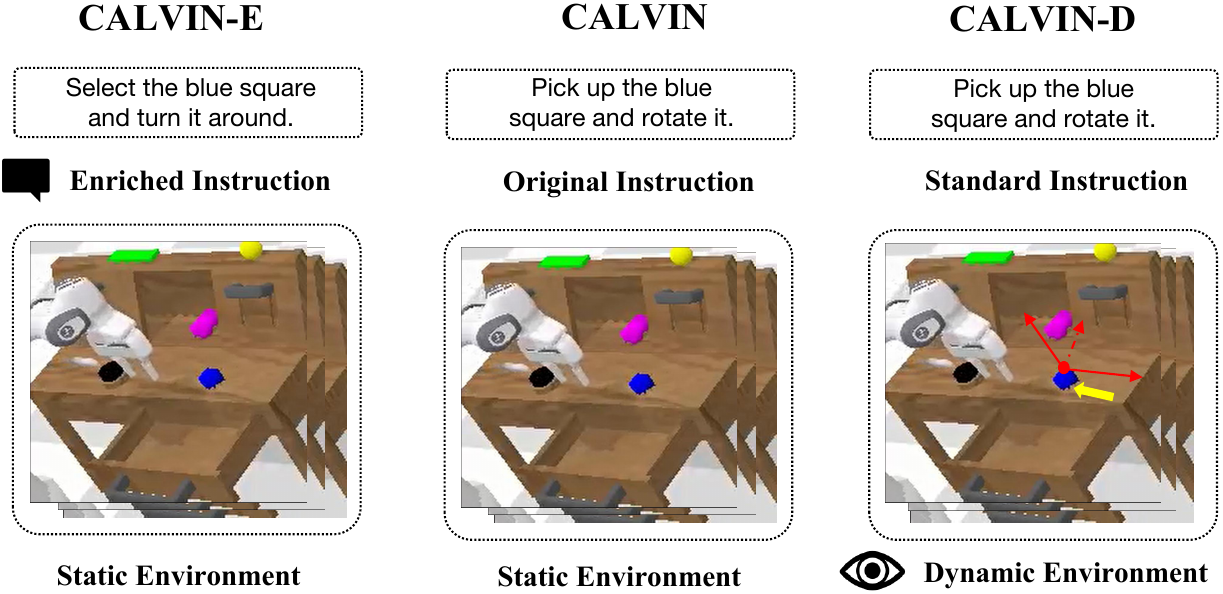} 
\end{center}
\caption{\textbf{Three Different Evaluation Environments.}}
\label{newfig: evaluation}
\end{figure}

\noindent\textbf{Dataset Processing.}
Unlike LCB~\cite{shentu2024llms}, which constructs a chat-like response before the <ACT> token, we directly concatenate an <ACT> token after the instruction.
This approach was adopted because we have not yet implemented this functionality. Additionally, we observed that even without implementing this feature, performance remains satisfactory. We plan to address the processing of chat-like data in future work.

\noindent\textbf{Environment.}
For comparison with models that are not open-source but have published results, we selected the same environment as they used. To maintain consistency with LCB~\cite{shentu2024llms} and RoboDual~\cite{bu2024towards}, we selected the CALVIN environment as our core comparative simulation environment. Real-world environment experiments will be supplemented later.

\noindent\textbf{Standard Evaluation.}
Following prior works, we primarily validate our effectiveness in the ABC-D scenario. 
To ensure rapid assessment of experiments, we used the first 100 evaluations from the standard 1000 evaluations, improving the testing efficiency for ablation experiments. 
In the final evaluation Table~\ref{tab:calvin_all}, we extend the evaluations to a full set of 1,000 to provide more comprehensive experimental results.

\noindent\textbf{Harder Evaluation.}
As is shown in Figure~\ref{newfig: evaluation}, in the standard evaluation test scenario, objects are static, and the given language instructions are standard.
However, the dual system should inherently combine the language generalization capabilities of large models with the advantages of small models' high-frequency characteristics for dynamic scenarios. Therefore, we conducted additional validation in two scenarios.

1. \textbf{CALVIN-E}: For language instruction generalization, we used Enriched language instructions for testing.

2. \textbf{CALVIN-D}: For dynamic scenario testing, in grasping tasks, we made objects move in four different ways within the environment to examine the model's robustness in dynamic scenarios.

\begin{table}[t]
    \centering
    \small
    \caption{\textbf{Evaluation of single system in CALVIN-D environment. } 
    }
    \vspace{-3pt}
    \scalebox{1.0}{
    \begin{tabular}{c|ccccc}
    \toprule
    \multirow{2}{*}{Model}  &\multicolumn{5}{c}{ Success rate (\%) $\uparrow$}  \\
    & Static & Left & Forward & Diagonal & Circle  \\
    \midrule     
    RF & 100 & 0 & 0 & 0 & 0 \\
    3DDA & 82 & 84 & 46 & 67 & 80 \\
    \bottomrule
    \end{tabular}
    }
    \label{tab: 1}
\end{table}

\begin{table}[t]
    \centering
    \small
    \caption{\textbf{Comparison of different training strategies for the low-level policy in standard CALVIN environment.} 
    }
    \vspace{-3pt}
    \scalebox{1.0}{
    \begin{tabular}{c|ccccc|c}
    \toprule
      \multirow{2}{*}{MLLM}  &\multicolumn{5}{c|}{Task completed in a row (\%) $\uparrow$} & \multirow{2}{*}{\makecell[c]{Avg. Len} $\uparrow$}  \\
      & 1 & 2 & 3 & 4 & 5 &  \\
    \midrule     
     Fine-tuning & 96 & 83 & 68 & 58 & 48 & 3.53 \\
      From-scratch & 89 & 71 & 49 & 42 & 34 & 2.85  \\
    \bottomrule
    \end{tabular}
    }
    \label{tab: which policy}
\end{table}

\begin{figure*}[t]
\begin{center}
\vspace{-1.2em}
  \includegraphics[width=1.0\textwidth]{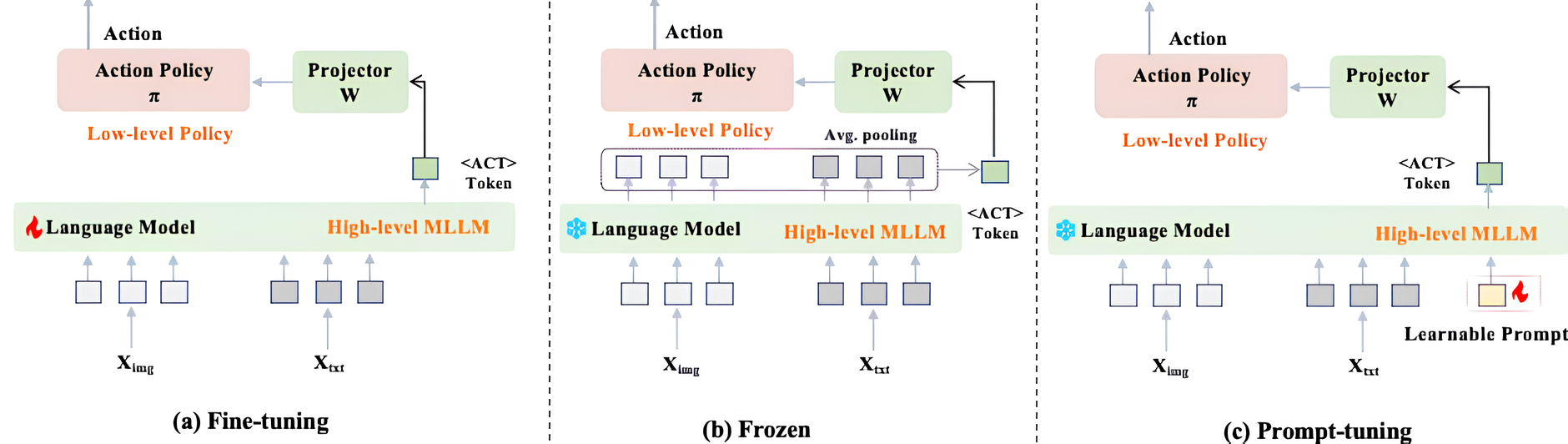} 
\end{center}
\caption{\textbf{Three Different MLLM Training Strategy.}}
\label{newfig: comparasion of MLLM}
\end{figure*}

\begin{table*}[t]
    \centering
    \small
    % \vspace{-12pt}
    \caption{\textbf{Comparison of different training strategies for the high-level MLLM in CALVIN environment.} 
    }
    \vspace{-3pt}
    \scalebox{1.0}{
    \begin{tabular}{c|c|c|c|ccccc|c}
    \toprule
    \multirow{2}{*}{Benchmark} & \multirow{2}{*}{MLLM} & \multirow{2}{*}{Integration of MLLM and Policy } &\multirow{2}{*}{Policy}  &\multicolumn{5}{c|}{Task completed in a row (\%) $\uparrow$} & \multirow{2}{*}{\makecell[c]{Avg. Len} $\uparrow$}  \\
     & & & &  1 & 2 & 3 & 4 & 5 &  \\
    \midrule     
    \multirow{4}{*}{CALVIN}& Frozen & w CLIP Loss & Fine-tuning & 94 & 80 & 64 & 51 & 41 & 3.30 \\
    & Frozen & w/o CLIP Loss & Fine-tuning & 90 & 74 & 61 & 54 & 40 & 3.33 \\
    & Fine-tuning & w CLIP Loss & Fine-tuning & 96 & 83 & 68 & 58 & 48 & 3.53  \\
    & Fine-tuning & w/o CLIP Loss & Fine-tuning & 88 & 72 & 56 & 46 & 30 & 3.13  \\
    \bottomrule
    \end{tabular}
    }
    \label{tab: which MLLM}
\end{table*}

\begin{table*}[t]
    \centering
    \small
    % \vspace{-12pt}
    \caption{\textbf{Further MLLM training experiments in CALVIN and CALVIN-E environment.} 
    }
    \vspace{-3pt}
    \scalebox{1.0}{
    \begin{tabular}{c|c|c|c|ccccc|c}
    \toprule
     \multirow{2}{*}{Benchmark} & \multirow{2}{*}{MLLM} & \multirow{2}{*}{Integration of MLLM and Policy } &\multirow{2}{*}{Policy}  &\multicolumn{5}{c|}{Task completed in a row (\%) $\uparrow$} & \multirow{2}{*}{\makecell[c]{Avg. Len} $\uparrow$}  \\
     & & & & 1 & 2 & 3 & 4 & 5 &  \\
    \midrule     
    \multirow{2}{*}{CALVIN} &Prompt-tuning & w CLIP Loss & Fine-tuning & 94 & 78 & 62 & 52 & 42 & 3.28  \\
    & Prompt-tuning & w/o CLIP Loss & Fine-tuning & 94 & 77 & 67 & 60 & 47 & 3.45  \\
    \midrule 
    \multirow{4}{*}{CALVIN-E} &Prompt-tuning & w CLIP Loss & Fine-tuning & 81 & 54 & 41 & 27 & 15 & 2.09  \\
     &Prompt-tuning & w/o CLIP Loss & Fine-tuning & 72 & 55 & 40 & 26 & 20 & 2.13  \\
    & Fine-tuning & w CLIP Loss & Fine-tuning & 76 & 49 & 30 & 15 & 4 & 1.74  \\
    & Frozen & w CLIP Loss & Fine-tuning & 72 & 37 & 21 & 11 & 5 & 1.46  \\
    \bottomrule
    \end{tabular}
    }
    \label{tab: which MLLM_v2}
\end{table*}

\subsection{Why not single system?}
\noindent\textbf{Preliminary.} In fact, the definition of dual systems has always been ambiguous, but since we established the CALVIN-D experiment, we found that previous single-system work (e.g., Roboflmanigo~\cite{li2023vision}) would directly fail under such testing, therefore subsequent experiments were not conducted on single systems.

\noindent\textbf{Setup.} The specific experimental configuration involved testing models trained on the standard ABC dataset on CALVIN-D for 100 trials. The "Static" condition represents scenarios where standard objects do not move, while "Left," "Forward," "Diagonal," and "Circle" represent four different object movement patterns. The specific results are shown in Table~\ref{tab: 1}.

\noindent\textbf{Analysis.} We discovered that the results of the RF~\cite{li2023vision} model on CALVIN-D were quite surprising, as it completely failed to complete the corresponding tasks in dynamic scenarios.
The primary reason for the observed performance is that, during the testing phase, the RF method requires processing the previous six image frames to obtain the corresponding latent representations for LSTM-based action inference. While the latent representations typically remain stable during training, they become variable in the testing phase as a result of object movement within dynamic scenarios. This discrepancy between training and testing conditions leads to a significant drop in performance, resulting in a consistently zero success rate in dynamic environments.
Nevertheless, we can also observe that the RF model using MLLM demonstrates extremely high performance on simple tasks, showing much greater robustness than the smaller 3DDA model. This highlights the significance of using MLLM as the "brain" of the system.

\noindent\textbf{Discussion.} 
Of course, we acknowledge that this conclusion may not be completely rigorous, as further testing on $\pi_0$~\cite{black2024pi_0}, GR00TN1~\cite{bjorck2025gr00t} has not yet been conducted. These additional experiments will be included in future work.

\subsection{Training strategy of dual system}
For dual-system models, the main training strategy consists of three parts: how to train the low-level policy, how to train the high-level MLLM, and how to connect the two. The following experiments will be divided into these three components.

\subsubsection{Policy Training Strategy.}
\noindent\textbf{Preliminary.} For LCB, the downstream low-level policy uses pre-trained 3DDA, while HiRT employs the RT-1 structure and trains from scratch. Robodual uses its own designed downstream policy. Setting aside the differences in configurations, there are two paradigms for policy training: \textbf{training from scratch} and \textbf{fine-tuning from pre-trained models}.

\noindent\textbf{Setup.} 
For fair comparison, the large model configuration follows the LCB structure: LLaVA1.0 backbone, connected with the <ACT> token, all using CLIP Loss to align the <ACT> token with downstream instructions. The only difference is that the downstream policy uses either a pre-trained 3DDA policy or a policy trained from scratch. The specific results are shown in Table~\ref{tab: which MLLM}.

\noindent\textbf{Analysis.} 
In Table~\ref{tab: which policy}, we discover that using a pre-trained policy can improve performance while reducing overall training time. Therefore, subsequent experiments are all based on \textbf{fine-tuning from pre-trained policy model}.

\subsubsection{MLLM Training Strategy.}
\noindent\textbf{Preliminary.} For LCB, HiRT, and Robodual, the upstream large models all underwent fine-tuning. Although GR00TN1~\cite{bjorck2025gr00t} doesn't fall within the scope of dual systems, it achieved excellent results by adopting a frozen paradigm for training. Therefore, we conducted experiments on both approaches.

\noindent\textbf{Setup.} 
For fair comparison, the large model configuration follows the LCB structure: LLaVA1.0 backbone, connected with the <ACT> token, all using CLIP Loss to align the <ACT> token with downstream instructions.
The downstream policy adopts the fine-tuning paradigm throughout. During the connection process between the MLLM and the policy model, we also introduced whether to include CLIP loss as a variable.

\noindent\textbf{Analysis.} 
For scenarios where the MLLM is frozen, adding or omitting the CLIP loss does not significantly affect performance. This is because the CLIP loss itself is meant to adjust the unchanged MLLM's output to accommodate the downstream small model's input, resulting in minimal performance differences. 
However, when the MLLM requires fine-tuning, the impact of CLIP loss becomes highly significant. Without the constraint of CLIP loss, it's easy to disrupt the small model's already-trained attention mechanisms between conditioning and other perceptual inputs, potentially leading to performance degradation. 

\begin{table*}[ht]
    \centering
    \small
    % \vspace{-12pt}
    \caption{\textbf{Performance of different projector initialization in CALVIN environment. Here, Pre-alignment refers to training the projector prior to training the MLLM.} 
    }
    \vspace{-3pt}
    \scalebox{1}{
    \begin{tabular}{c|c|c|c|c|ccccc}
    \toprule
    \multirow{2}{*}{Benchmark} &\multirow{2}{*}{Pre-alignment}& \multirow{2}{*}{MLLM} & \multirow{2}{*}{Integration of MLLM and Policy} &\multirow{2}{*}{Policy}  &\multicolumn{5}{c}{Task completed in a row (\%) $\uparrow$} \\
     & & & & & 1 & 2 & 3 & 4 & 5   \\
    \midrule     
    \multirow{6}{*}{CALVIN} & \checkmark & Frozen & w CLIP Loss & Fine-tuning & 94 & 80 & 64 & 51 & 41  \\
    & \checkmark & Fine-tuning   & w CLIP Loss & Fine-tuning & 96 & 83 & 68 & 58 & 48   \\
    & \checkmark & Prompt-tuning   & w/o CLIP Loss & Fine-tuning & 94 & 77 & 67 & 60 & 47   \\
    & $\times$ & Frozen  &w CLIP Loss & Fine-tuning & 0 & 0 & 0 & 0 & 0 \\
    & $\times$ & Fine-tuning   & w CLIP Loss & Fine-tuning & 0 & 0 & 0 & 0 & 0 \\
    & $\times$ & Prompt-tuning    & w/o CLIP Loss & Fine-tuning & 0 & 0 & 0 & 0 & 0  \\
 
    \bottomrule
    \end{tabular}
    }
    \label{tab: which projector}
\end{table*}

\noindent\textbf{Intuitive hypothesis.} 
Although the introduction of CLIP loss makes the overall model performance functional, this approach essentially compromises the large model's inherent generalization capabilities. Is there a way to keep the large model parameters frozen while still ensuring that the large model can be updated together with the downstream components?

\noindent\textbf{Further setup.}
As shown in Figure~\ref{newfig: comparasion of MLLM}, we only changed the training method of the MLLM
Specifically, we adopted prompt tuning. We added a new <ACT> token to the large model's vocabulary and only trained the lm-head layer while keeping all other model parameters fixed. This approach essentially trains an additional token in the vocabulary that only relates to downstream tasks, without altering the MLLM model's inherent generalization capabilities. Therefore, theoretically, it can better ensure the connection between the dual systems. 

Next, we use experiments to verify this hypothesis in Table~\ref{tab: which MLLM_v2}.

\noindent\textbf{Further analysis.}
For the prompt tuning paradigm, while performance in the standard Calvin testing environment is comparable to other training paradigms, there are significant differences in experiments validating language generalization.
Similarly, under the premise of having CLIP loss, the generalization capability of prompt tuning results far exceeds that of fine-tuning and frozen approaches. Moreover, without CLIP loss supervision, generalization actually improves somewhat, which fully demonstrates that the prompt tuning paradigm trains the large model with minimal dependence on altering the large model's generalization capabilities.

\begin{figure*}[ht]
\begin{center}
\vspace{-.2em}
  \includegraphics[width=0.85\textwidth]{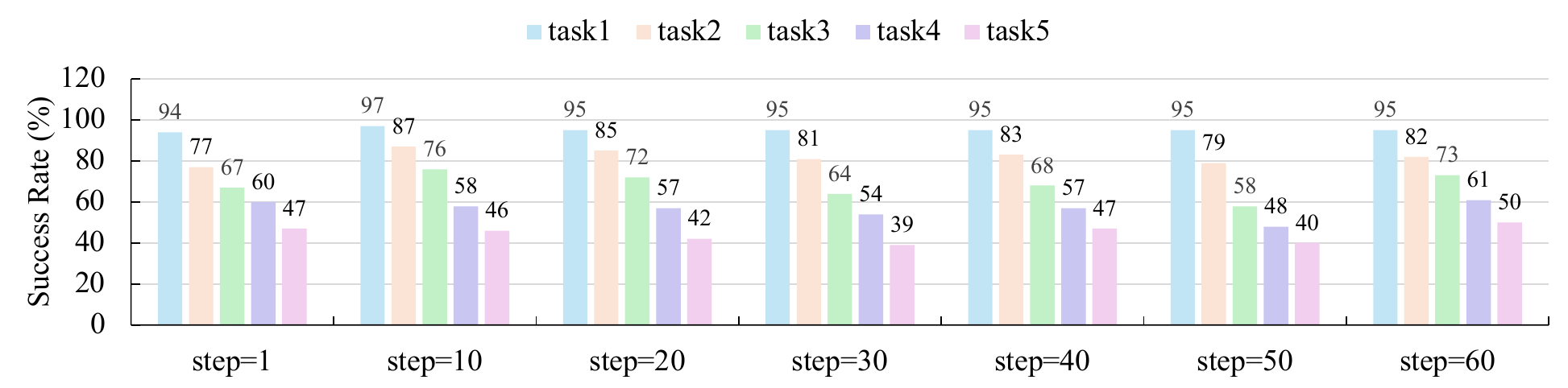} 
\end{center}
\vspace{-1.2em}
\caption{
\textbf{Evaluations on hierarchical inference.} We evaluate the performance of the dual system on the CALVIN benchmark, with inference steps set to 1 and 60, respectively.``Steps" refers to the inference steps of action policy during a single MLLM inference step. The longest environmental steps of the action policy~\cite{ke20243d} are 60, which means MLLM only inference once and represents the most typical asynchronous scenarios.
}
\label{tab:asynchronous steps}
% \vspace{-1.5em}
\end{figure*}
\begin{figure*}[ht]
\begin{center}
% \vspace{-.2em}
  \includegraphics[width=\textwidth]{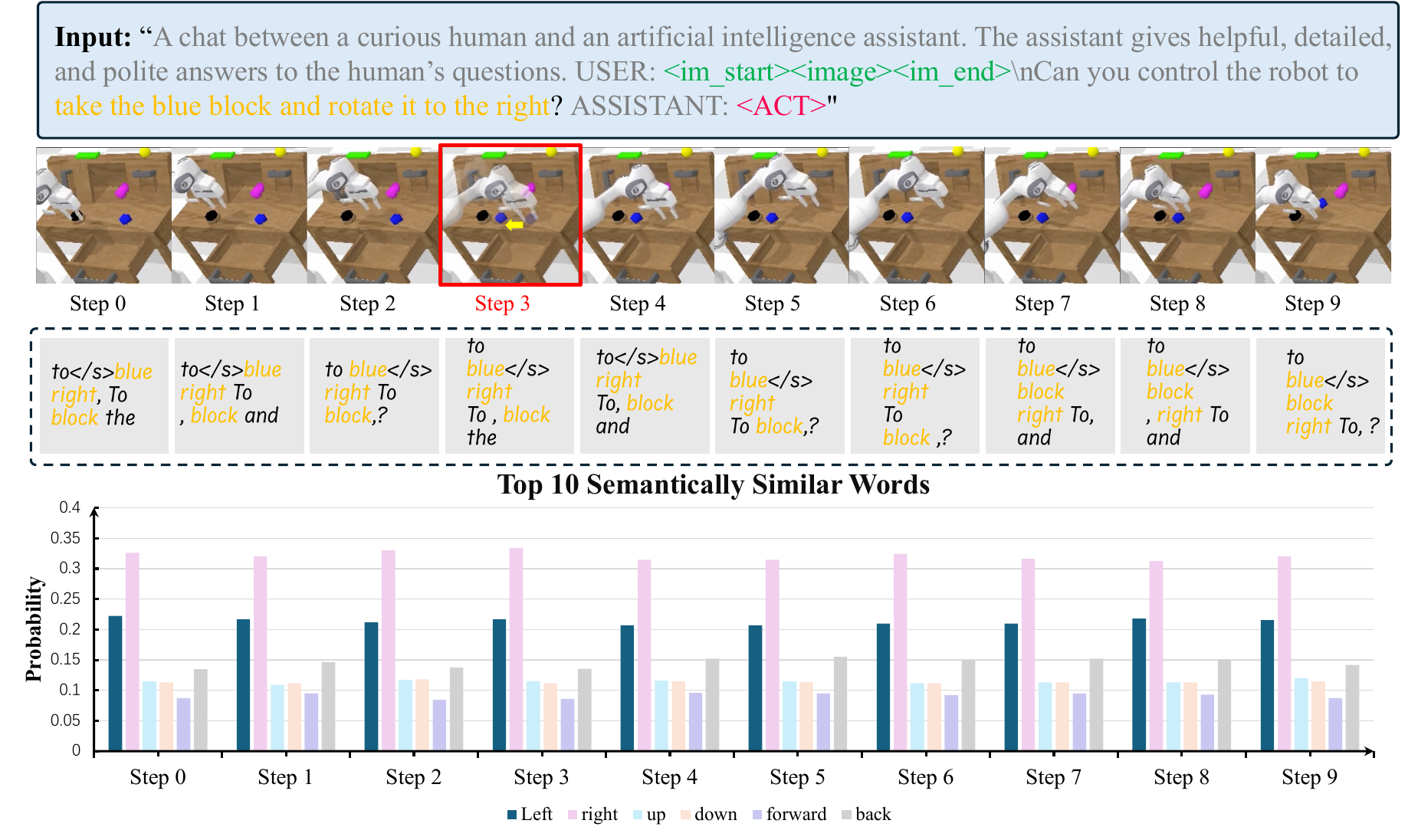} 
\end{center}
% \vspace{-1.5em}
\caption{\textbf{Evaluation on the shortcoming of existing dual systems.} From top to bottom, the first row displays the input to the MLLM. The second row visualizes a special scenario where, at environment step 3, the blue block is manually shifted to the left. In the third row, we present the top 10 words that are semantically closest to the latent embedding. The bottom row illustrates the probability distribution of spatial words associated with the latent embedding.}
\label{fig:visual_reasoning}
% \vspace{-0.2em}
\end{figure*} 

\subsubsection{Dual-System Integration Strategy.}

\noindent\textbf{Preliminary.}

Based on the experiments above, our conclusion is that using a pre-trained policy and fine-tuning the MLLM with prompt tuning yields the best results. However, this still involves the process of how to connect the components, as the semantic gap between upstream and downstream can be substantial. Therefore, we primarily conduct the following ablation analysis.

\noindent\textbf{Setup.}
% 用于衔接上下游需要依赖于一个MLP projector 去实现，我们在这里采用了两种方式去实现：第一种将上下游模型直接解冻，将上下游模型和MLP projector 一起联合训练，第二种方式：先将上游大模型冻住，训练MLP projector和下游小模型，然后再将上游大模型解冻，一起训练。这两种方式之间的差异主要在于是否有一个MLP projector的过程。
To connect the upstream and downstream components, an MLP projector is needed. We implemented two approaches here: First, directly unfreezing both upstream and downstream models and jointly training them together with the MLP projector. Second, initially freezing the upstream large model while training the MLP projector and downstream small model, then unfreezing the upstream large model for joint training. The main difference between these two approaches is whether there is a separate MLP projector training process. The result is in Table~\ref{tab: which projector}.

\noindent\textbf{Analysis.}
We found that without prior projector pre-alignment, connecting upstream and downstream models based on frozen, fine-tuning and prompt tuning approaches directly fails. This demonstrates the importance of Projector pre-alignment in the connection process.
Of course, if we adopt a train-from-scratch approach for the downstream policy, a two-stage process is not required. However, as shown in Table~\ref{tab: 1}, training from scratch produces inferior results.

\subsection{Testing strategy of dual system}
\noindent\textbf{Preliminary.} A key essence of dual-system models is the need to implement asynchronous control between upper and lower layers. In LCB, the authors did not specifically handle asynchronous operations, instead using synchronous training followed by asynchronous inference. In HiRT, the authors adopted an additional buffer to introduce asynchronous operations during training as well. For Robodual, they utilized real-time replacement of the upper layer's coarse actions with actions inferred by the lower layer to perform asynchronous operations. Here, we primarily validated the first approach, with the latter two paradigms to be updated subsequently.

\noindent\textbf{Setup.} We evaluated different asynchronous steps from 1 to 60 on CALVIN-D. The step refers to the inference steps of action policy during a single MLLM inference step. The
longest environmental steps of the 3DDA are 60. 

\noindent\textbf{Analysis.} 
We observe a surprising conclusion in Figure~\ref{tab:asynchronous steps}: regardless of the number of steps between the large model's inferences, the performance changes are quite similar.
Moreover, even in dynamic scenarios, the experimental results are consistent. 

\noindent\textbf{Intuitive hypothesis.} 
This result indicates that the current MLLM is not sensitive to changes in the current environment, which is counterintuitive. 
Therefore, we need to clarify exactly what information is being transmitted from the upper layer's latent vector to the lower layer.

\begin{table*}[ht]
    \centering
    \small
    % \vspace{-12pt}
    \caption{\textbf{Comparison of the using method of MLLM.} 
    }
    \vspace{-3pt}
    \scalebox{1.0}{
    \begin{tabular}{c|c|c|c|ccccc|c}
    \toprule
    \multirow{2}{*}{Benchmark} & \multirow{2}{*}{Type of MLLM} & \multirow{2}{*}{Auxiliary tasks} &\multirow{2}{*}{Policy}  &\multicolumn{5}{c|}{Task completed in a row (\%) $\uparrow$} & \multirow{2}{*}{\makecell[c]{Avg. Len} $\uparrow$}  \\
     & & & & 1 & 2 & 3 & 4 & 5 &  \\
    \midrule     
    \multirow{3}{*}{CALVIN}& MLLM (Prompt Tuning) & $\times$ & Fine-tuning & 94 & 77 & 67 & 60 & 47 & 3.45 \\
    & LLM (Prompt Tuning) & $\times$& Fine-tuning & 77 & 48 & 26 & 16 & 10 & 1.77 \\
    & MLLM (Prompt Tuning) & \checkmark & Fine-tuning & 98 & 92 & 76 & 72 & 63 & 4.01  \\
    \bottomrule
    \end{tabular}
    }
    \label{tab: what extend}
\end{table*}

\noindent\textbf{Further setup.} 
To explore the underlying reasons, we mapped the latent embeddings of action tokens into semantic space and calculated the similarity of different words to analyze what these action tokens from MLLM convey. The experiment involves dynamic scenarios where a blue block consistently moves to the left. The result is in Figure~\ref{fig:visual_reasoning}.

\noindent\textbf{Further analysis.} We have the following conclusions:

1. As to the similarity with spatial words at different time steps, we observe that regardless of whether the robotic arm moves left or right, the probability of ``right" is consistently higher than that of ``left," while the probabilities of different spatial prepositions remain almost unchanged over time. This indicates that the action token has learned a semantic feature that remains constant and is unrelated to changes in the environment. The higher probability of ``right" compared to ``left" may be due to ``right" carrying more semantic information; for example, ``right" can also imply correctness, contributing to its consistently high probability.

2. As to Top 10 similar words at different time steps, we observe that the latent embedding primarily encodes the target object, spatial relations, and action semantics from the instruction, along with some noise. It means that the latent embedding mainly summarizes the textual instruction and is largely insensitive to changes in visual information.
In other words, the current training method does not effectively leverage the visual reasoning capabilities of the MLLM. Instead, the MLLM merely transmits the semantics of the instructions to the low-level policy.

\subsection{Whether the MLLM of dual system is enough?}

\noindent\textbf{Preliminary.}
Based on the experimental analysis above, the information currently transmitted through the latent token is insufficient for the downstream model to effectively complete tasks. Therefore, in this section, we aim to explore better ways to utilize upstream information.

\noindent\textbf{Setup.}
The experiments are based on the above conclusions, with the downstream model using fine-tuning, adopting a two-stage projector training approach, and the upstream large model utilizing a prompt tuning training paradigm. However, three variants were created for how to use MLLM: 1. Standard MLLM; 2. Removing visual information from MLLM, treating it purely as an LLM; 3. Introducing an auxiliary loss, allowing the generated latent token to connect to an additional head layer to infer action-related information(position or rotation). The result is in Table~\ref{tab: what extend}.

\noindent\textbf{Analysis.}
From the experimental results, it can be seen that using only LLM produces results far inferior to MLLM, which demonstrates the inherent function of MLLM and shows it hasn't degraded to simply functioning as an LLM. When we have additional auxiliary tasks, we can see a significant increase in the success rate of tasks. This is mainly because the additional auxiliary tasks force the model to capture more visual information in order to accomplish them, thus compelling the model to pay attention to tasks that a purely MLLM approach would not focus on.

\begin{figure}[t]
    \centering
    \includegraphics[width=8 cm]{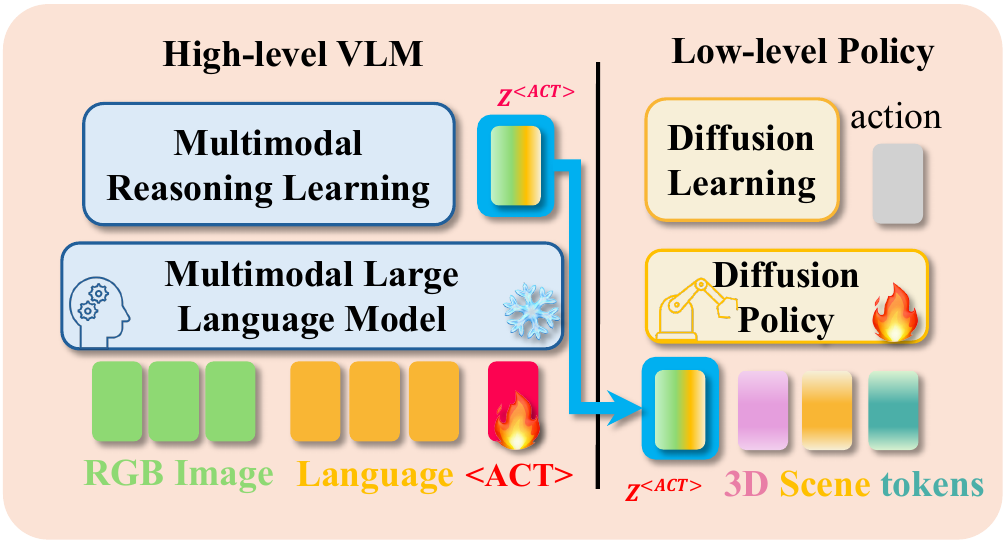}
    \caption{\textbf{Overview of our proposed Dual System VLA.}}
    \label{fig:teaser}
\end{figure}

\begin{figure*}[ht]
\begin{center}
% \vspace{-.2em}
  \includegraphics[width=0.85\textwidth]{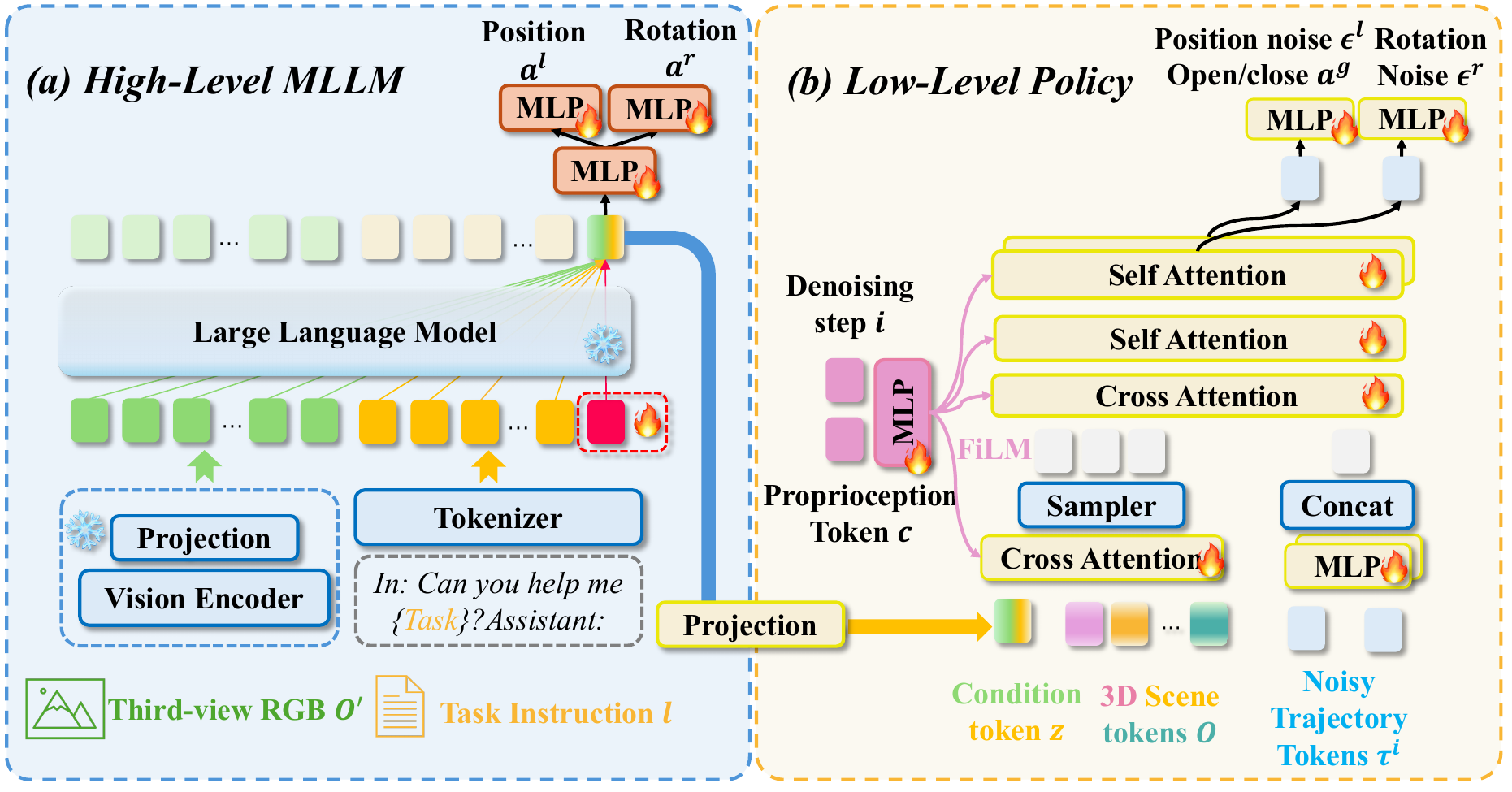} 
\end{center}
% \vspace{-1.2em}
\caption{
\textbf{Detailed framework.} (a) The high-level MLLM (left) takes third-view RGB $o'$, task instruction $l$, and a learnable token $\act$ as input. After processing through the Large Language Model (LLM), we extract the feature embedding from the final layer of the $\act$ token as the latent goal for the low-level policy. To fully leverage the MLLM’s multimodal reasoning capability, we propose an auxiliary task, using MLPs to predict the action (location $a^l$, rotation $a^r$, open/close $a^g$) based on this feature embedding $z^{\act}$, ensuring it encapsulates both visual and textual information. (b) The low-level policy (right) receives the latent goal from the high-level MLLM, combines it with 3D scene tokens $o$ and proprioception token $c$, and iteratively predicts action noise $\epsilon$ to produce an accurate action trajectory $\tau$ and gripper state $a^g$. Notably, our approach keep all parameters of the MLLM frozen and fine-tune the learnable prompt to adjust the MLLM’s output, significantly reducing training costs compared to previous methods.
}
\label{fig:framework}
% \vspace{-1.5em}
\end{figure*}

\section{A Simple yet Effective Dual System VLA}

Based on the above analysis, we employ prompt-tuning to adapt the output of the large model rather than directly fine-tuning the MLLM itself.
Additionally, we introduce an auxiliary task to exploit MLLM's visual reasoning capabilities fully.
This approach results in a more robust latent embedding that effectively integrates visual and textual information.

\subsection{Architecture}

\noindent\textbf{Network.}
Our system comprises two main components: a pre-trained MLLM $f_{\phi}$ and a pre-trained policy $\pi_{\theta}$, with parameters $\phi$ and $\theta$, respectively. The MLLM includes a text-only large language model and a vision encoder, which projects images into the embedding space of the language model, allowing for a multi-modal understanding of textual and visual inputs. The pre-trained policy consists of a vision encoder and transformer-based diffusion model. Using multiple cross-attention layers, the diffusion model incorporates a lot of conditioning information, such as 3D scene representations, proprioception information, and condition/instruction tokens from the high-level model. In this work, we leverage LLaVA [27] as the high-level MLLM and 3D Diffuser Actor as the low-level pre-trained diffusion policy. Notably, we use a linear layer to replace the 3D Diffuser Actor's text encoder, aligning the dimension of the latent embedding output by the large model with the input dimension of the low-level policy.

\noindent\textbf{Input and Output.} The whole system is designed to mimic demonstration trajectories in the format $\{l,(o_1,a_1),(o_2,a_2),...\}$, where $l=\{w_i \in \mathrm{R}^d\}_{i=1}^N$ represents a task-specific language instruction of length N with an input dimension d, and $o_t$ and $a_t$ denote the visual observation and corresponding robot action at each timestep $t$. The input observation $o_t$ consists of two RGB-D images from different viewpoints. The output action $a_t$ defines the end-effector’s pose, which is decomposed into 3D location, rotation, and gripper state (open/close): $a_t= \{a_t^{\rm l}\in \mathbb{R}^3,a_t^{\rm r}\in \mathbb{R}^6,a_t^{\rm g}\in \{0,1\}\}$.
The MLLM $f_{\phi}$ processes language instruction $l$ and the third-view RGB image $o'_t$, outputting the latent embedding $z_t$ for low-level policy. The low-level pre-trained policy $\pi_{\theta}$ takes as input the noisy trajectory $\tau_t^i$, diffusion step $i$, and the conditioning information from the environment observation $o_t$, the latent embedding $z_t$, and proprioception $c_t$ of timestep $t$, predicting the action trajectory $\tau_t = (a_{t:t+T}^{\rm l},a_{t:t+T}^{\rm r})$ and binary states $a_{t:t+T}^{g}$ at each timestep $t$, over a temporal horizon $T$. 

\begin{table*}[!t]
    \centering
    \caption{ 
    Results on CALVIN ABC-D: We report both success rates and average task completion length (out of 5 tasks) per evaluation sequence. MLLM (PT) denotes our proposed prompt tuning method for MLLM training. Policy(P) indicates loading from a pretrained policy model. Asy(10) represents inference with a 10-step time delay. AUX denotes the additionally introduced auxiliary tasks.}
    \vspace{-4pt}
    \label{tab:calvin_all}
    \small
    \setlength{\tabcolsep}{8pt}
    \scalebox{1.0}{
    \begin{tabular}{l|c|l|ccccc|c}
    \toprule
    \multicolumn{1}{l|}{\multirow{2}{*}{Type}}& \multicolumn{2}{l|}{\multirow{2}{*}{Method}}   & \multicolumn{5}{c|}{ Task completed in a row (\%) $\uparrow$} & \multirow{2}{*}{Avg. Len. $\uparrow$} \\
    \multicolumn{1}{l|}{} & \multicolumn{2}{l|}{}  & 1 & 2 & 3 & 4 & 5 & \\
    \midrule
    % \multicolumn{2}{l|}{3D Diffusion Policy} & Lang & 28.7 & 2.7 & 0.0 & 0.0 & 0.0 & 0.31 \\
    \multirow{4}{*}{CALVIN}& \multicolumn{2}{l|}{Only Policy}  & 92.2 & 78.7 & 63.9 & 51.2 & 41.2 & 3.27 \\
    & \multicolumn{2}{l|}{MLLM (PT) + Policy(P)}  & 92.2 & 79.2 & 65.0 & 52.9 & 40.9 & 3.30 \\
    & \multicolumn{2}{l|}{MLLM (PT) + AUX + Policy(P) + Asy(10)}  & \textbf{93.3} & \textbf{81.8} & \textbf{67.9} & \underline{56.6} & \underline{46.0} & \textbf{3.45} \\
    & \multicolumn{2}{l|}{MLLM (PT) + AUX + Policy(P) + Asy(60)}  & \underline{92.8} & \underline{79.7} & \underline{67.5} & \textbf{57.3} & \textbf{46.9} & \underline{3.44} \\
    \midrule
    \multirow{4}{*}{CALVIN-E}    &\multicolumn{2}{l|}{Only Policy}  & 65.2 & 39.1 & 20.3 & 11.7 & 6.1 & 1.42 \\
    & \multicolumn{2}{l|}{MLLM (PT) + Policy(P)}  & 71.3 & 44.9 & 28.4 & 17.5 & 10.3 & 1.72 \\
    & \multicolumn{2}{l|}{MLLM (PT) + AUX + Policy(P) + Asy(10)}  & \textbf{78.9} & \textbf{57.1} & \textbf{40.2} & \textbf{29.5} & \textbf{20.2} & \textbf{2.26} \\
    & \multicolumn{2}{l|}{MLLM (PT) + AUX + Policy(P) + Asy(60)}  & \underline{78.1} & \underline{56.5} & \underline{38.9} & \underline{27.0} & \underline{19.5} & \underline{2.20} \\
    \midrule
    \end{tabular}
    }
    \vspace{-3pt}
\end{table*}

\subsection{Training}
\noindent\textbf{Prompt Tuning.} In order to avoid the degradation of MLLM, we introduce one learnable token $\act \in \mathbb{R}^d$ at end of language instruction $l$. The new instruction $l'$ is defined as $l'=\{l,\act\}$. During training, all parameters of MLLM are frozen; we only update the embedding of learnable token \act.

\noindent\textbf{Multimodal Reasoning Learning.}
As we discussed in section 3.3, we know that these previous methods do not fully utilize MLLM's visual reasoning capability. Specifically, they align the output of the large MLLM model with the output from the text encoder of CLIP. Using purely textual information to supervise the fine-tuning of the MLLM can lead to the degradation of multimodal reasoning capability. Therefore, we designed an auxiliary task to leverage the multimodal reasoning capability of the MLLM fully. 
This task is very simple and requires no additional data preparation process. The output embedding $z_t^\act=f_{\phi}(o'_t,l')$ from the learnable prompt token is passed through linear layers to predict the action trajectories $\tau_t$ and gripper actions $a^{g}_t$. Through supervised training on this task, we ensure that the large model has to utilize visual input information and that the latent embedding contains a blend of multimodal information. The loss function is defined as follows:
\begin{equation}
\begin{aligned}
    \mathcal{L}_{lm}(\act) &= {\rm BCE}({\rm MLP}(f_{\phi}^{g}(o'_t,l')),a_{t:t+T}^{g})\\ &+ \omega_1 \cdot ||{\rm MLP}(f_{\phi}^{l}(o'_t,l'))-a_{t:t+T}^{l}||\\ &+ \omega_2 \cdot ||{\rm MLP}(f_{\phi}^{r}(o'_t,l'))-a_{t:t+T}^{r}||,
\end{aligned}
\end{equation}
where $\omega_1$, $\omega_2$ are hyperparameters to balance the effect of each loss item, and $\rm MLP$ represents linear layer. To reconstruct the sequence of 3D locations and 3D rotations, we apply the $L_1$loss. Additionally, we supervise the end-effector opening using binary cross-entropy loss (BCE).

\noindent\textbf{Diffusion Learning.} Following the previous diffusion-based approach \cite{chi2023diffusion,ke20243d,ze20243d}, we train our model using the action denoising objective. During training, we randomly sample a time step $t$ and a diffusion step $i$, adding noise $\epsilon=(\epsilon^l,\epsilon^r)$ to a ground-truth trajectory $\tau_t^0$. The objective is defined as:
\begin{equation}
\begin{aligned}
    \mathcal{L}_{policy}(\theta,\act) &= {\rm BCE}(\pi_{\theta}^{g}(o_t,z_t^{\act},c_t,\tau_t^i,i),a_{t:t+T}^{g})\\ &+ \omega_3 \cdot ||\epsilon_{\theta}^{l}(o_t,z_t^{\act},c_t,\tau_t^i,i)-\epsilon_{t:t+T}^{l}||\\ &+ \omega_4 \cdot ||\epsilon_{\theta}^{r}(o_t,z_t^{\act},c_t,\tau_t^i,i)-\epsilon_{t:t+T}^{r}||,
\end{aligned}
\end{equation}
where $\omega_3$, $\omega_4$ are also hyperparameters to balance loss items. Please refer to [1] for the details of the loss function.

\noindent\textbf{Two stage training.}
We adopt a two-stage training approach to train our proposed dual system. In the first stage, to initially align the embedding produced by the MLLM with the feature space of the pre-trained policy, we freeze the parameters of the large model and the low-level policy, training only the prompt and projection layers. In the second stage, we keep the large model frozen and unfreeze the low-level policy, fine-tuning it together with the prompt and projection. The objectives in both stages remain unchanged. The only difference between the two stages is whether the low-level policy is frozen.
In summary, our loss function includes two components and can be defined as follows:
\begin{equation}
    \mathcal{L}_{total} = \mathcal{L}_{lm}+\mathcal{L}_{policy}
\end{equation}

\noindent\textbf{Implementation Details.}
We use LLaVA-7B\cite{llava} and 3D diffuser Actor\cite{ke20243d} as the high-level MLLM and low-level policy models, respectively. We select the checkpoint of 3D diffuser Actor at 65,000 iterations as the pre-trained parameters. During training, the first stage (pre-alignment) is conducted for 2,000 iterations, and the second stage continues until 100,000 iterations. The projection is a linear layer that reduces the output dimension of the large model from 4096 to 512. We manually add an \act token to LLava’s tokenizer and freeze all parameters of the MLLM, fine-tuning only the newly added token embedding. The remaining experimental and training settings are consistent with 3D diffuser Actor; please refer to \cite{ke20243d} for details.

\subsection{Results}

From Table~\ref{tab:calvin_all}, we can draw conclusions similar to those from the previous empirical study:

1. The crucial role of integrating the upper and lower layers of the model lies in improving performance in language generalization scenarios.

2. Additional auxiliary tasks are very helpful for enhancing both standard task performance and generalization performance, mainly because they improve the model's action capability.

3. Asynchronous inference has little impact on the inference performance of the general task model; even if the model only performs asynchronous inference once (Asy (60)), the final performance remains largely unchanged.

\section{Discussion \& Limitation}

We first acknowledge that there is still a long way to go before we achieve a full open-source reproduction of Helix.

1. Deploying on real robots.
2. Achieving sufficiently fast downstream policy execution.
3. Successfully running on physical robots.
4. Deploying on humanoid robots.
5. Realizing collaboration between humanoid robots.

There is indeed much work to be done before all of the above goals are achieved. However, this technical report is only our initial version. We are committed to continuously updating this project to fulfill the open-source objectives for all the tasks mentioned above.
In addition, we maintain an open attitude toward some of the claims in this article that have not yet been fully verified. We hope that more researchers will join our team, or that more people will support us, so that we can accomplish this meaningful work for the entire community. Since some of the organization was done rather hastily, if any corrections are needed, all authors are welcome to contact me at any time.

\clearpage
{
    \small
    \bibliographystyle{ieeenat_fullname}
    \bibliography{main}

\begin{thebibliography}{25}
\providecommand{\natexlab}[1]{#1}
\providecommand{\url}[1]{\texttt{#1}}
\expandafter\ifx\csname urlstyle\endcsname\relax
  \providecommand{\doi}[1]{doi: #1}\else
  \providecommand{\doi}{doi: \begingroup \urlstyle{rm}\Url}\fi

\bibitem[Belkhale and Sadigh(2024)]{belkhale2024minivla}
S Belkhale and D Sadigh.
\newblock Minivla: A better vla with a smaller footprint.
\newblock 2024.

\bibitem[Bjorck et~al.(2025)Bjorck, Casta{\~n}eda, Cherniadev, Da, Ding, Fan, Fang, Fox, Hu, Huang, et~al.]{bjorck2025gr00t}
Johan Bjorck, Fernando Casta{\~n}eda, Nikita Cherniadev, Xingye Da, Runyu Ding, Linxi Fan, Yu Fang, Dieter Fox, Fengyuan Hu, Spencer Huang, et~al.
\newblock Gr00t n1: An open foundation model for generalist humanoid robots.
\newblock \emph{arXiv preprint arXiv:2503.14734}, 2025.

\bibitem[Black et~al.(2024)Black, Brown, Driess, Esmail, Equi, Finn, Fusai, Groom, Hausman, Ichter, et~al.]{black2024pi_0}
Kevin Black, Noah Brown, Danny Driess, Adnan Esmail, Michael Equi, Chelsea Finn, Niccolo Fusai, Lachy Groom, Karol Hausman, Brian Ichter, et~al.
\newblock A vision-language-action flow model for general robot control.
\newblock \emph{arXiv preprint arXiv:2410.24164}, 2024.

\bibitem[Brohan et~al.(2023)Brohan, Brown, Carbajal, Chebotar, Dabis, Finn, Gopalakrishnan, Hausman, Herzog, Hsu, Ibarz, Ichter, Irpan, Jackson, Jesmonth, Joshi, Julian, Kalashnikov, Kuang, Leal, Lee, Levine, Lu, Malla, Manjunath, Mordatch, Nachum, Parada, Peralta, Perez, Pertsch, Quiambao, Rao, Ryoo, Salazar, Sanketi, Sayed, Singh, Sontakke, Stone, Tan, Tran, Vanhoucke, Vega, Vuong, Xia, Xiao, Xu, Xu, Yu, and Zitkovich]{brohan2023rt1roboticstransformerrealworld}
Anthony Brohan, Noah Brown, Justice Carbajal, Yevgen Chebotar, Joseph Dabis, Chelsea Finn, Keerthana Gopalakrishnan, Karol Hausman, Alex Herzog, Jasmine Hsu, Julian Ibarz, Brian Ichter, Alex Irpan, Tomas Jackson, Sally Jesmonth, Nikhil~J Joshi, Ryan Julian, Dmitry Kalashnikov, Yuheng Kuang, Isabel Leal, Kuang-Huei Lee, Sergey Levine, Yao Lu, Utsav Malla, Deeksha Manjunath, Igor Mordatch, Ofir Nachum, Carolina Parada, Jodilyn Peralta, Emily Perez, Karl Pertsch, Jornell Quiambao, Kanishka Rao, Michael Ryoo, Grecia Salazar, Pannag Sanketi, Kevin Sayed, Jaspiar Singh, Sumedh Sontakke, Austin Stone, Clayton Tan, Huong Tran, Vincent Vanhoucke, Steve Vega, Quan Vuong, Fei Xia, Ted Xiao, Peng Xu, Sichun Xu, Tianhe Yu, and Brianna Zitkovich.
\newblock Rt-1: Robotics transformer for real-world control at scale, 2023.

\bibitem[Bu et~al.(2024)Bu, Li, Chen, Cai, Zeng, Cui, Yao, and Qiao]{bu2024towards}
Qingwen Bu, Hongyang Li, Li Chen, Jisong Cai, Jia Zeng, Heming Cui, Maoqing Yao, and Yu Qiao.
\newblock Towards synergistic, generalized, and efficient dual-system for robotic manipulation.
\newblock \emph{arXiv preprint arXiv:2410.08001}, 2024.

\bibitem[Chi et~al.(2023)Chi, Xu, Feng, Cousineau, Du, Burchfiel, Tedrake, and Song]{chi2023diffusion}
Cheng Chi, Zhenjia Xu, Siyuan Feng, Eric Cousineau, Yilun Du, Benjamin Burchfiel, Russ Tedrake, and Shuran Song.
\newblock Diffusion policy: Visuomotor policy learning via action diffusion.
\newblock \emph{The International Journal of Robotics Research}, page 02783649241273668, 2023.

\bibitem[Evans(2008)]{evans2008dual}
Jonathan St~BT Evans.
\newblock Dual-processing accounts of reasoning, judgment, and social cognition.
\newblock \emph{Annu. Rev. Psychol.}, 59\penalty0 (1):\penalty0 255--278, 2008.

\bibitem[Gong et~al.(2024)Gong, Ding, Lyu, Huang, Sun, Zhao, Fan, and Wang]{gong2024carp}
Zhefei Gong, Pengxiang Ding, Shangke Lyu, Siteng Huang, Mingyang Sun, Wei Zhao, Zhaoxin Fan, and Donglin Wang.
\newblock Carp: Visuomotor policy learning via coarse-to-fine autoregressive prediction.
\newblock \emph{arXiv preprint arXiv:2412.06782}, 2024.

\bibitem[Han et~al.(2024)Han, Kim, and Jang]{han2024dual}
ByungOk Han, Jaehong Kim, and Jinhyeok Jang.
\newblock A dual process vla: Efficient robotic manipulation leveraging vlm.
\newblock In \emph{Conference on Robot Learning (CoRL)}, 2024.

\bibitem[Kahneman(2011)]{kahneman2011thinking}
Daniel Kahneman.
\newblock \emph{Thinking, fast and slow}.
\newblock macmillan, 2011.

\bibitem[Ke et~al.(2024)Ke, Gkanatsios, and Fragkiadaki]{ke20243d}
Tsung-Wei Ke, Nikolaos Gkanatsios, and Katerina Fragkiadaki.
\newblock 3d diffuser actor: Policy diffusion with 3d scene representations.
\newblock \emph{arXiv preprint arXiv:2402.10885}, 2024.

\bibitem[Lai et~al.(2024)Lai, Dai, Chen, Pang, Rehg, and Liu]{lai2024lego}
Bolin Lai, Xiaoliang Dai, Lawrence Chen, Guan Pang, James~M Rehg, and Miao Liu.
\newblock Lego: L earning ego centric action frame generation via visual instruction tuning.
\newblock In \emph{European Conference on Computer Vision}, pages 135--155. Springer, 2024.

\bibitem[Li et~al.(2023)Li, Liu, Zhang, Yu, Xu, Wu, Cheang, Jing, Zhang, Liu, et~al.]{li2023vision}
Xinghang Li, Minghuan Liu, Hanbo Zhang, Cunjun Yu, Jie Xu, Hongtao Wu, Chilam Cheang, Ya Jing, Weinan Zhang, Huaping Liu, et~al.
\newblock Vision-language foundation models as effective robot imitators.
\newblock \emph{arXiv preprint arXiv:2311.01378}, 2023.

\bibitem[Liu et~al.(2023{\natexlab{a}})Liu, Li, Wu, and Lee]{liu2023visual}
Haotian Liu, Chunyuan Li, Qingyang Wu, and Yong~Jae Lee.
\newblock Visual instruction tuning.
\newblock \emph{arXiv preprint arXiv:2304.08485}, 2023{\natexlab{a}}.

\bibitem[Liu et~al.(2023{\natexlab{b}})Liu, Li, Wu, and Lee]{llava}
Haotian Liu, Chunyuan Li, Qingyang Wu, and Yong~Jae Lee.
\newblock Visual instruction tuning.
\newblock In \emph{Thirty-seventh Conference on Neural Information Processing Systems}, 2023{\natexlab{b}}.

\bibitem[Neys(2006)]{neys2006dual}
Wim~De Neys.
\newblock Dual processing in reasoning: Two systems but one reasoner.
\newblock \emph{Psychological science}, 17\penalty0 (5):\penalty0 428--433, 2006.

\bibitem[Pan et~al.(2025)Pan, Shukla, Singh, Zhao, Mishra, Wang, Xu, Chen, Li, Juefei-Xu, et~al.]{pan2025transfer}
Xichen Pan, Satya~Narayan Shukla, Aashu Singh, Zhuokai Zhao, Shlok~Kumar Mishra, Jialiang Wang, Zhiyang Xu, Jiuhai Chen, Kunpeng Li, Felix Juefei-Xu, et~al.
\newblock Transfer between modalities with metaqueries.
\newblock \emph{arXiv preprint arXiv:2504.06256}, 2025.

\bibitem[Reuss et~al.()Reuss, Zhou, R{\"u}hle, Ya{\u{g}}murlu, Otto, and Lioutikov]{reussflower}
Moritz Reuss, Hongyi Zhou, Marcel R{\"u}hle, {\"O}mer~Erdin{\c{c}} Ya{\u{g}}murlu, Fabian Otto, and Rudolf Lioutikov.
\newblock Flower: Democratizing generalist robot policies with efficient vision-language-action flow policies.
\newblock In \emph{7th Robot Learning Workshop: Towards Robots with Human-Level Abilities}.

\bibitem[Shentu et~al.(2024)Shentu, Wu, Rajeswaran, and Abbeel]{shentu2024llms}
Yide Shentu, Philipp Wu, Aravind Rajeswaran, and Pieter Abbeel.
\newblock From llms to actions: Latent codes as bridges in hierarchical robot control.
\newblock \emph{arXiv preprint arXiv:2405.04798}, 2024.

\bibitem[Su et~al.(2025)Su, Zhan, Fang, Xue, Fang, Li, Lu, and Yang]{su2025dense}
Yue Su, Xinyu Zhan, Hongjie Fang, Han Xue, Hao-Shu Fang, Yong-Lu Li, Cewu Lu, and Lixin Yang.
\newblock Dense policy: Bidirectional autoregressive learning of actions.
\newblock \emph{arXiv preprint arXiv:2503.13217}, 2025.

\bibitem[Tversky and Kahneman(1974)]{tversky1974judgment}
Amos Tversky and Daniel Kahneman.
\newblock Judgment under uncertainty: Heuristics and biases: Biases in judgments reveal some heuristics of thinking under uncertainty.
\newblock \emph{science}, 185\penalty0 (4157):\penalty0 1124--1131, 1974.

\bibitem[Wen et~al.(2025)Wen, Zhu, Li, Tang, Shen, and Feng]{wen2025dexvla}
Junjie Wen, Yichen Zhu, Jinming Li, Zhibin Tang, Chaomin Shen, and Feifei Feng.
\newblock Dexvla: Vision-language model with plug-in diffusion expert for general robot control.
\newblock \emph{arXiv preprint arXiv:2502.05855}, 2025.

\bibitem[Ze et~al.(2024)Ze, Zhang, Zhang, Hu, Wang, and Xu]{ze20243d}
Yanjie Ze, Gu Zhang, Kangning Zhang, Chenyuan Hu, Muhan Wang, and Huazhe Xu.
\newblock 3d diffusion policy: Generalizable visuomotor policy learning via simple 3d representations.
\newblock In \emph{ICRA 2024 Workshop on 3D Visual Representations for Robot Manipulation}, 2024.

\bibitem[Zhang et~al.(2024)Zhang, Guo, Chen, Wang, Hu, Shi, and Chen]{zhang2024hirt}
Jianke Zhang, Yanjiang Guo, Xiaoyu Chen, Yen-Jen Wang, Yucheng Hu, Chengming Shi, and Jianyu Chen.
\newblock Hirt: Enhancing robotic control with hierarchical robot transformers.
\newblock \emph{arXiv preprint arXiv:2410.05273}, 2024.

\bibitem[Zhu et~al.(2024)Zhu, Zhu, Li, Wen, Xu, Liu, Cheng, Shen, Peng, Feng, et~al.]{zhu2024scaling}
Minjie Zhu, Yichen Zhu, Jinming Li, Junjie Wen, Zhiyuan Xu, Ning Liu, Ran Cheng, Chaomin Shen, Yaxin Peng, Feifei Feng, et~al.
\newblock Scaling diffusion policy in transformer to 1 billion parameters for robotic manipulation.
\newblock \emph{arXiv preprint arXiv:2409.14411}, 2024.

\end{thebibliography}
}

\end{document}